\documentclass[10pt,journal,compsoc]{IEEEtran}

\ifCLASSOPTIONcompsoc
  \usepackage[nocompress]{cite}
\else
  \usepackage{cite}
\fi

\ifCLASSINFOpdf
   \usepackage[pdftex]{graphicx}
\else
   \usepackage[dvips]{graphicx}
\fi

\usepackage{amsmath}

\usepackage{algorithmic}

\usepackage{array}

\ifCLASSOPTIONcompsoc
 \usepackage[caption=false,font=footnotesize,labelfont=sf,textfont=sf]{subfig}
\else
 \usepackage[caption=false,font=footnotesize]{subfig}
\fi

\begin{document}
\title{Learning Spatial and Temporal Variations for 4D Point Cloud Segmentation}

\author{Shi~Hanyu,
        Wei~Jiacheng, Wang~Hao,
        Liu~Fayao and Lin~Guosheng%
\IEEEcompsocitemizethanks{\IEEEcompsocthanksitem Shi H., Wei J., Wang H. and Lin G. are with Nanyang Technological University\protect\\
E-mail: \{hanyu001, jiacheng002, hao005\}@e.ntu.edu.sg, gslin@ntu.edu.sg
\IEEEcompsocthanksitem Liu F. is with Institute for Infocomm Research A*STAR, Singapore.\protect\\
E-mail: fayaoliu@gmail.com
}%
\thanks{Manuscript received August 19, 2021.}}%

\ifCLASSOPTIONpeerreview

    \markboth{Journal of \LaTeX\ Class Files,~Vol.~14, No.~8, August~2015}%
    {Shell \MakeLowercase{\textit{et al.}}: Bare Demo of IEEEtran.cls for Computer Society Journals}
\fi

\IEEEtitleabstractindextext{%
\begin{abstract}
LiDAR-based 3D scene perception is a fundamental and important task for autonomous driving. 
Most state-of-the-art methods on LiDAR-based 3D recognition tasks focus on single frame 3D point cloud data, and the temporal information is ignored in those methods. We argue that the temporal information across the frames provides crucial knowledge for 3D scene perceptions, especially in the driving scenario. 
In this paper, we focus on spatial and temporal variations to better explore the temporal information across the 3D frames.
We design a temporal variation-aware interpolation module and a temporal voxel-point refiner to capture the temporal variation in the 4D point cloud. 
The temporal variation-aware interpolation generates local features from the previous and current frames by capturing spatial coherence and temporal variation information.
The temporal voxel-point refiner builds a temporal graph on the 3D point cloud sequences and captures the temporal variation with a graph convolution module.
The temporal voxel-point refiner also transforms the coarse voxel-level predictions into fine point-level predictions. 
With our proposed modules, the new network TVSN achieves state-of-the-art performance on SemanticKITTI and SemantiPOSS. Specifically, our method achieves 52.5\% in mIoU (+5.5\%  against previous best approaches) on the multiple scan segmentation task on SemanticKITTI, and 63.0\% on SemanticPOSS (+2.8\% against previous best approaches).
\end{abstract}

}

\maketitle

\IEEEdisplaynontitleabstractindextext

\ifCLASSOPTIONpeerreview
\begin{center} \bfseries EDICS Category: 3-BBND \end{center}
\fi
\IEEEpeerreviewmaketitle

\IEEEraisesectionheading{\section{Introduction}\label{sec:introduction}}

\IEEEPARstart{L}{iDAR} 3D point clouds contain a series of points to represent the surrounding information. 
Nowadays, the sensing accuracy of LiDAR devices fulfils the requirement of real-world applications. 
Compared to 2D images, 3D point cloud data delivers far more accurate depth/distance information.
The advantages attract many researchers into the area of 3D point cloud scene perception. As a fundamental task of 3D scene perception, semantic segmentation on the 3D point cloud predicts the labels of each point in the point cloud.

\begin{figure}
    \begin{center}
	\includegraphics[width=1.0\linewidth]{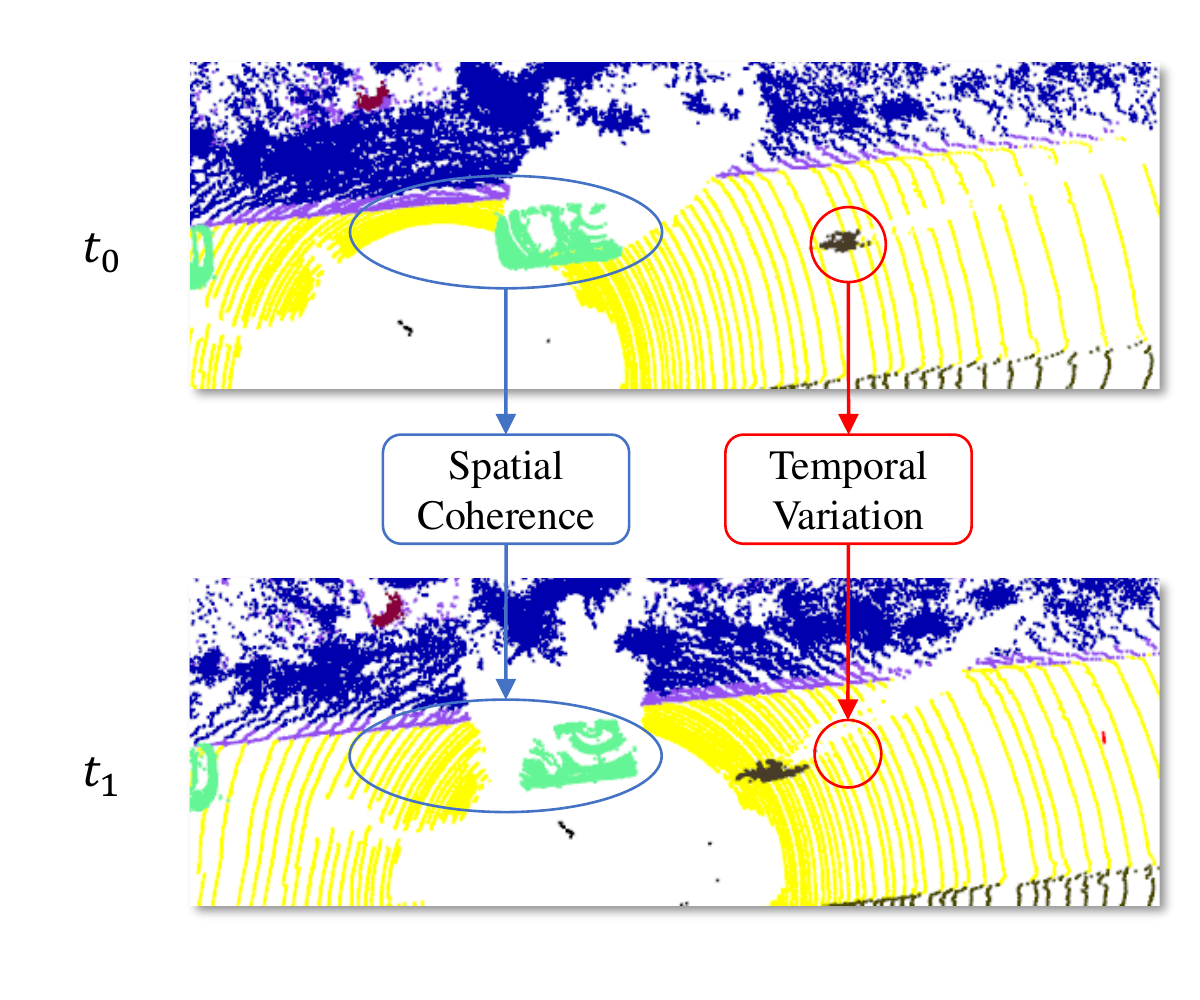}  
    \end{center}

	\caption{\textbf{A visualization of temporal information.} We illustrate two corresponding point cloud frames in the same location. In blue area, the spatial coherence represents the similarity of the area in two frames. In red area, the temporal variation represents the differences in two frames. 
	}
	\label{fig:seq}
\end{figure}

\begin{figure*}
    \begin{center}
	\subfloat[LiDAR 3D point cloud]{
			\includegraphics[width=1.0\linewidth]{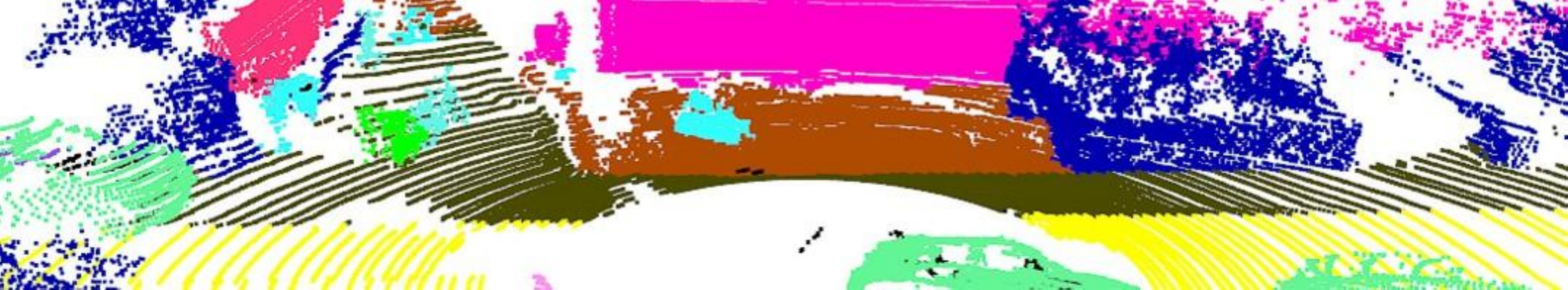}  
			\label{fig:ex_p}
		}
    \end{center}
    \begin{center}
	\subfloat[Voxelized 3D point cloud]{
			\includegraphics[width=1.0\linewidth]{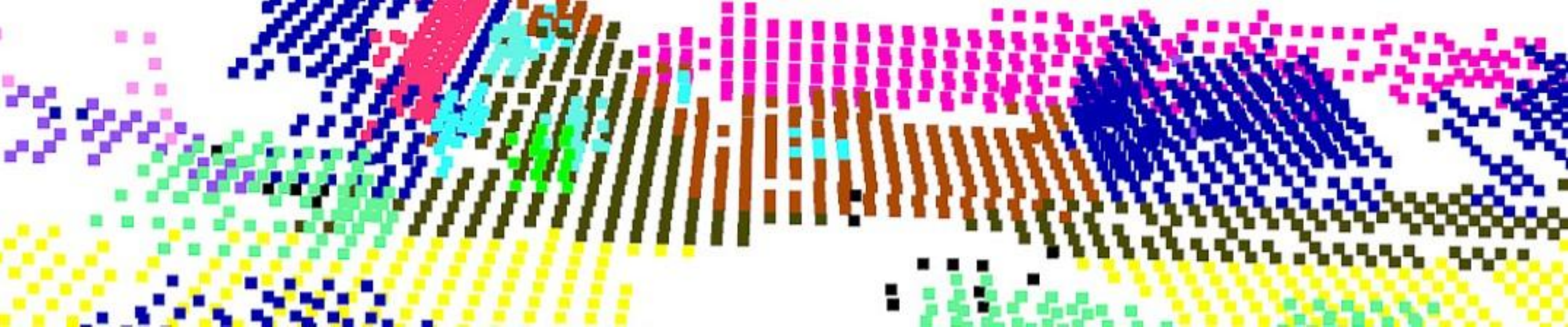}  
			\label{fig:ex_v}
		}
    \end{center}    

	\caption{\textbf{A comparison between different process method.} All three examples are selected in the same region from the same frame.
	}
	\label{fig:com_pvp}
\end{figure*}

A typical 3D point cloud semantic segmentation model predicts the label of each point in a single frame. Single frame based methods solely rely on geometric information and other features within the frame. To improve the performance of single frame methods, the recent approaches like RPVNet~\cite{xu2021rpvnet} and AF2-S3Net~\cite{cheng20212} adopt very large models and require very high computational resources, which may indicate that the cost to improve point cloud segmentation performance is quite high with only spatial information from a single frame.
In real world applications, LiDAR point cloud data is collected as a sequence of point cloud frames, which can be considered as 4D point clouds. 
In a 4D point cloud sequence, temporal information plays a crucial role in scene perception, especially in complex outdoor scenes. 
Besides, processing 4D point cloud sequences is closer to human behavior since we are constantly receiving data with our eyes. Our brain will also take previous frames to help us understand the scene for the current moment.
Comparing to 3D point cloud semantic segmentation, there are two main advantages in 4D point cloud semantic segmentation. 
\begin{itemize}
    \item Firstly, multi-frame point clouds sequences contains temporal information compared to single frame point cloud data. 
By utilizing both temporal and spatial information, we can produce more accurate predictions and handle some complex tasks. 
    \item Secondly, segmentation models can produce more robust predictions with temporal information from neighboring frames, and it encourages consistent predictions across frames.
\end{itemize}
Accordingly, 4D point cloud semantic segmentation is a promising research area in scene perception.

However, the performance of existing point cloud sequence methods~\cite{shi2020spsequencenet, duerr2020lidar} is barely satisfying.
We rethink the relationship between the spatial and temporal information in the 3D scenario to improve the existing methods. 
To extract temporal information, existing 3D point cloud sequence works~\cite{puy2020flot,shi2020spsequencenet,liu2019meteornet} utilize the similarity of local features in corresponding frames, 
which is 
called spatial coherence.
However, only utilizing the spatial coherence limits the performance of 4D point cloud segmentation methods.
As shown in Figure~\ref{fig:seq}, when there are some moving objects in the region, the features of the moving objects and the region on the object trajectory may change dramatically in the consecutive frames.
We define the changes of local regions in consecutive frames as temporal variation. 
The temporal variation reflects the motion information for both objects and environments. 

To this end, we propose two modules, i.e., temporal variation-aware interpolation and temporal voxel-point refiner, to exploit spatial coherence and temporal variation. 
Firstly, the temporal variation-aware interpolation is designed to exploit spatial coherence and temporal variation in local regions and perform the interpolation.
This module improves the network performance, especially on moving objects.
Secondly, we propose a simple yet effective temporal voxel-point refiner to capture detailed temporal information on the point level.
In this module, to associate the local features across the corresponding frames, we build a directed temporal graph on each point of the corresponding frames. 
We also design a temporal graph convolution to capture the temporal information and propagate the information on the temporal graph. 
Furthermore, as shown in Figure~\ref{fig:com_pvp}, the geometric details in each voxel are missing in the volumetric-based structure. 
Temporal voxel-point refiner rebuilds the missing details with the geometric details from the previous frame. 
We design a two-stage training pipeline to approach our network with high efficiency.

In summary, our contributions are:
\begin{itemize}
    \item We propose a novel spatial temporal point cloud segmentation network for efficient and robust segmentation by utilizing temporal information across frames.

    \item We design a temporal variation-aware interpolation module to accurately fuse the information from previous frames with the current frame by capturing the spatial coherence and temporal variation of local regions in neighbouring frames. 

    \item We propose a lightweight temporal voxel-point refiner to capture fine-level temporal information by building a temporal graph across frames and recover the details lost during data voxelization. In our design, the temporal voxel-point refiner improves the performance of 4D point cloud segmentation with a low computation cost.

    \item Our method achieves a significant improvement and outperforms all previous methods on SemanticKITTI~\cite{behley2019semantickitti} and SemanticPOSS~\cite{pan2020semanticposs}. In particular, our method achieves an mIOU score of 52.5\% (+5.5\% with previous best approaches) on the test set of SemanticKITTI~\cite{behley2019semantickitti} and 63.0\% on SemanticPOSS.

\end{itemize}

A preliminary version of our paper, which is denoted as SpSequenceNet, is published in ~\cite{shi2020spsequencenet}. 
Compared to SpSequenceNet, we have made three key improvements:
\begin{enumerate}
    \item We propose the temporal variation-aware interpolation module, an improved version of the cross-frame temporal interpolation in SpSequenceNet. 
    The temporal variation-aware interpolation captures the local differences of the features between the previous frame and the current frame to improve feature quality.

    \item We propose a temporal voxel-point refiner to dig in more temporal information and generate point level detailed features. 
    The temporal voxel-point refiner builds a temporal graph on the 4D point clouds and propagates the information along the time dimension with a graph convolution.  
    The temporal voxel-point refiner also transforms the voxel-level features to the point level features to recover the missing details during the voxelization. Our temporal voxel-point refiner improves the performance with a low computation requirement.
    
    \item We improve the backbone network and design a new training pipeline for our proposed methods. The new backbone network and training pipeline significantly outperform our old version. 
\end{enumerate}

\section{Related Work}

\subsection{3D Point Cloud Segmentation}
Point clouds contain rich environment information, which can benefit the 3D semantic segmentation for the robotic systems.
With the increase of dimension and huge amount of points, high quality feature extraction and low computation requirement are the keys in the 3D point cloud semantic segmentation. Therefore, many recent works achieve great improvements on both the precision and efficiency of 3D point cloud segmentation.
In this section, we discuss three types of methods, projection-based methods, volumetric-based methods and point-based methods.

\noindent \textbf{Projection-based Method.} Projection-based methods process the huge amount of points with the real-time requirement. 
Projection-based methods project the 3D point cloud data onto one or more 2d planes. For example, in multi-view approaches~\cite{lawin2017deep,boulch2017unstructured}, point clouds are projected onto several images from different views. In spherical approaches~\cite{wu2018squeezeseg,wu2019squeezesegv2}, each point cloud is projected onto one spherical plane. Then, they apply a 2D network on the projected plane. As stated in Section~\ref{sec:introduction}, the distortion of objects and the fuzzy of the projections limits the performance of these projection-based method, which is also the key area in the recent works~\cite{xu2020squeezesegv3,cortinhal2020salsanext}. 

\noindent \textbf{Volumetric-based methods.} Volumetric-based methods are extensions of dense convolution to reduce the computation requirement for the sparse 3D point cloud. 
A common way is to discretize the coordinates of points and index the discrete coordinates with a tree-based index~\cite{su2018splatnet,graham2014spatially,riegler2017octnet} or a hash table~\cite{graham2017submanifold,choy20194d}. Then, the convolutions are only applied on the transformed points based on the offset and the index. With a good choice of index algorithm, volumetric-based methods can reach a good performance with an acceptable speed. Nevertheless, the discretization highly affects the performance of volumetric-based methods, and also leads to some deviations during the inference step. To solve this problem, AF2-S3Net~\cite{cheng20212} extracts the point features with an additional encoder and decoder modules. In SPVNAS~\cite{tang2020searching}, they also design a sparse point-voxel convolution to fuse the point features and voxel features. 

\noindent \textbf{Point-based methods.} Inspired by Pointnet~\cite{qi2016pointnet}, there is a popular trend to adapt point-based method to 3D perception task, like 3D semantic segmentation and 3D scene flow. To capture the detail structure of point clouds, point-based methods apply a k nearest neighbours on every points. 
Then, there are three major methods to generate high quality features of local structures. 
Firstly, Pointnet-like methods~\cite{qi2017pointnetplusplus,jiang2018pointsift,zhang2019shellnet} design MLP structures to capture geometric structures with the colors and coordinates. Secondly, point convolution~\cite{li2018pointcnn,thomas2019kpconv,kochanov2020kprnet} design a convolution operation on the continuous coordinate system. Thirdly, graph-based methods~\cite{wang2019dynamic} set every points as nodes and the relationship between points as edges. Then, graph-based methods~\cite{wang2019dynamic} captures the local structure features with a graph neural network. In conclusion, point-based methods have good capabilities at capturing the local geometric features. However, due to the high computation requirement of nearest neighbor searching, point-based methods are hard to process large scale data. 

\begin{figure*}
    \centering
	\includegraphics[width=1.0\linewidth]{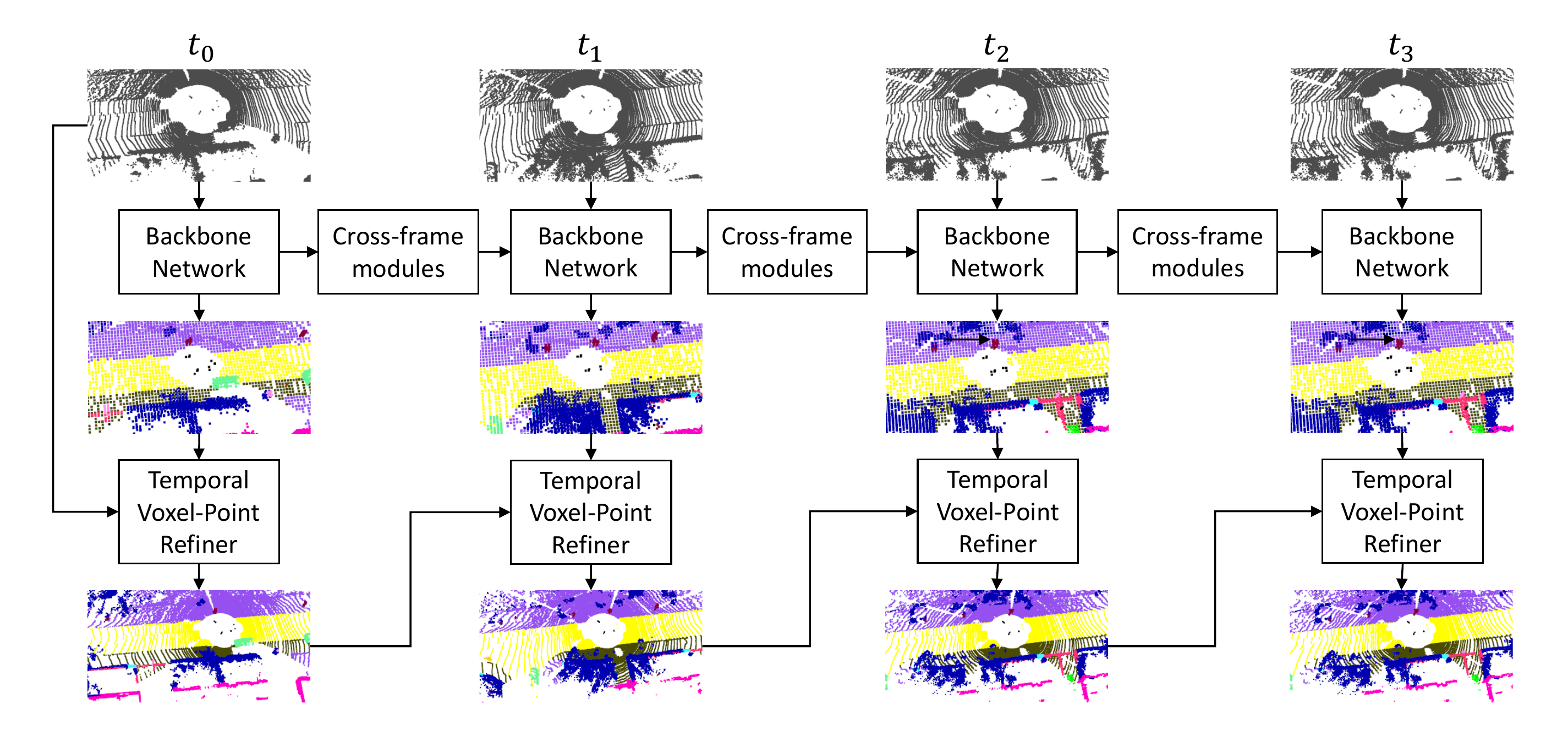}  

	\caption{\textbf{The structure of out proposed methods.} The backbone network is a 3D MinkowskiUNet. In the phase of encoder, there is two cross-frame modules, which are cross-frame global attention and temporal variation-aware interpolation. Then, we fuse the voxel predictions of the current backbone network and the point predictions from previous frame with a temporal voxel-point refiner.
	}
	\label{fig:network}

\end{figure*}

\subsection{4D Point Cloud Segmentation}

As a promising area for real-world applications, 4D point cloud segmentation analyzes the scene along the space and time dimensions. There are only a few methods working in 4D segmentations, and this area is yet under-investigated.

MinkowskiUNet~\cite{choy20194d} generalizes a formulation of sparse convolution in higher dimensions. They also investigate 4D sparse convolutions for 4D point cloud segmentation. As 4D sparse convolution requires high computation and memory costs, they propose a tesseract 4D convolution and achieves similar performance with lower computations. However, similar to the 3D convolution methods in 2D videos, the 4D convolution has efficiencies in 4D point cloud segmentation. 

Some other methods combine the features from multiple frames to extract the temporal information. SpSequenceNet~\cite{shi2020spsequencenet} adapts a 3D sparse unet and proposes two cross-frame modules to fuse the spatial and temporal information. 
The principle of cross-frame modules is the spatial coherence of two corresponding frames. 
TemporalLidarSeg~\cite{duerr2020lidar} design a projection-based structure to combine the information in a sequence of point clouds. 
Furthermore, they propose a ConvGRU to maintain a temporal memory. ConvGRU propagates the information from previous frames to the current inference step. 
ConvGRU relies on spatial coherence for propagation. 
However, similar to some assumptions~\cite{voigtlaender2019feelvos,oh2019video} in video object segmentation, the temporal information contains coherence and variation. 
We summarize the temporal information as spatial coherence and temporal variation.  
Only using spatial coherence limits the performance of previous works on the segmentation task. 
Furthermore, as stated in Section~\ref{sec:introduction}, there are some drawbacks in the backbone structures of both volumetric-based and spherical projection methods. 

Subsequently, we also explore more works in the video object segmentation. One recent novel work is space-time memory, STM~\cite{oh2019video}. 
STM proposes a memory mechanism to maintain the information of target objects from previous frames. For each iteration, STM applies an attention-based matching to search the pixels of the potential target object. 
However, the basic structure of STM has rooms for better performance and STM can not search multiple target objects. 
Hence, there are several improved methods~\cite{oh2019video,zhou2019enhanced, zhou2019motion, yang2020collaborative}. One method is FEELVOS~\cite{voigtlaender2019feelvos}. 
FEELVOS designs a local matching and a global matching to extract the object features with the local information. On the other hand, there are several differences between 3D point cloud segmentation and 2D video objects segmentation. Firstly, there are a few predefined objects in the first frame of 2D videos while 4D point cloud semantic segmentation is a general task without any additional information. 
Furthermore, the number of objects in a 4D point cloud is more than the number in a 2D video.
This difference reduces the value of global-wise matching, like the memory mechanism. 
Secondly, as the accuracy of point coordinates, compared to 2D video, 4D point clouds contain more spatial and temporal information and consume more computation racecourses, which increases the difficulties of balancing the performance and speed in the 4D point cloud segmentation methods.

\begin{figure*}
    \begin{center}
        \subfloat[Cross-frame Local Interpolation in our SpSequenceNet]{
			\includegraphics[width=0.75\linewidth]{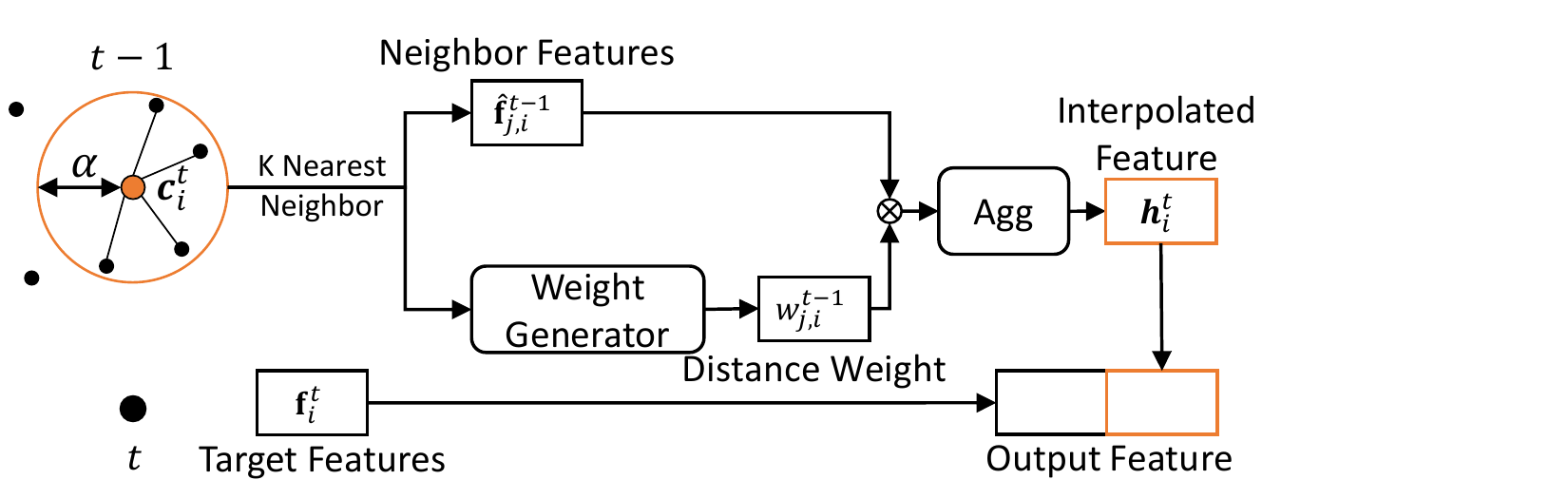}  
			\label{fig:CLI}
		}

    	\subfloat[temporal variation-aware interpolation]{
    			\includegraphics[width=0.75\linewidth]{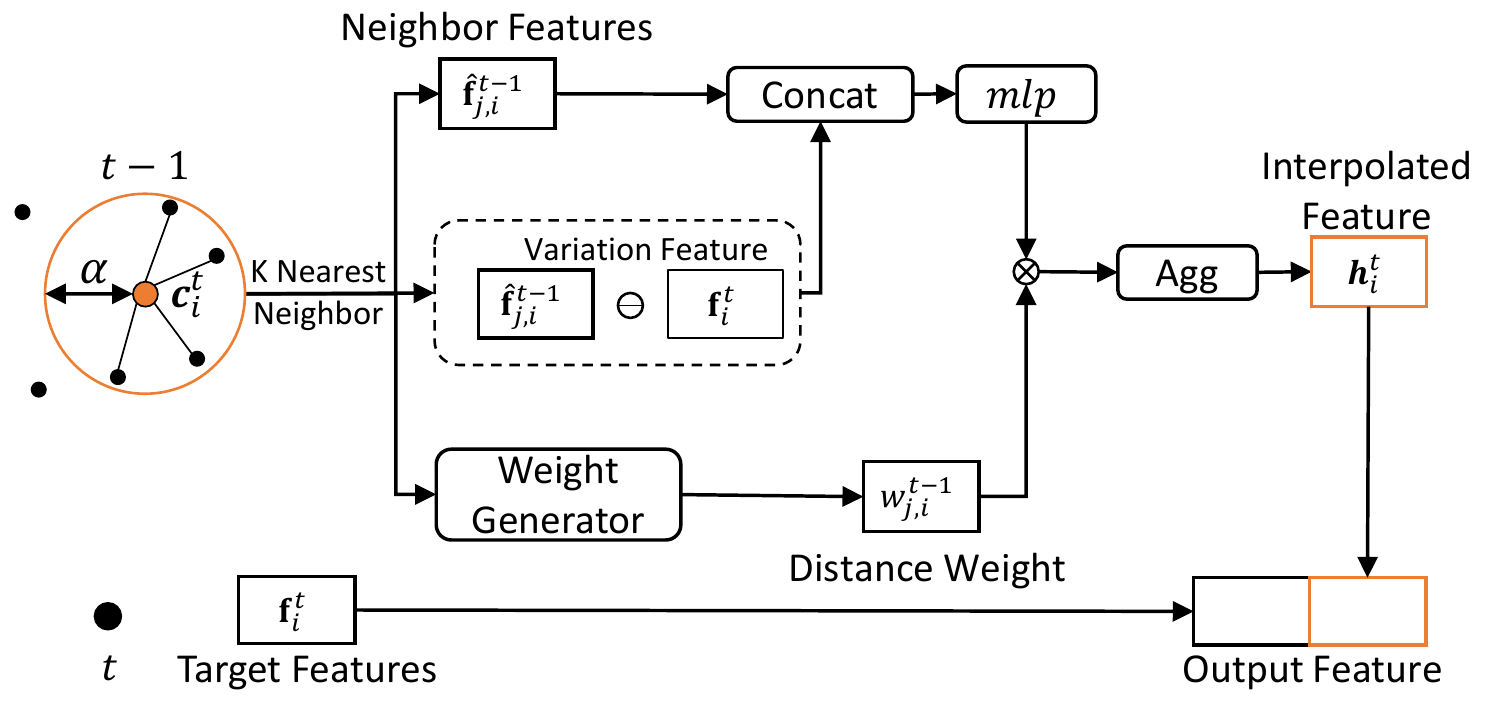}  
    			\label{fig:TVCLI}
    		}
    \end{center}
	
	\caption{\textbf{A comparison between cross-frame local interpolation and our proposed method.} The yellow point represents the coordinate $c_{i,t}$ to be interpolated in current frame while black points represent the coordinate in previous frame. $\alpha$ is the max distance in the neighbor search.  $\ominus$ is the element-wise minus. 
	}
	\label{fig:com_TVCLI}
	\vspace{-20pt}
\end{figure*}

\section{Temporal Variation Sequence Network}

In this section, we firstly revisit SpSequenceNet.
Then, we introduce our new temporal variation-aware interpolation and present a temporal voxel-point refiner to remap the voxelized prediction of the main network to the point-style prediction. We introduce the new backbone network and the training pipeline designed for our proposed modules.  We also illustrate the structure of our proposed network in Figure~\ref{fig:network}.

\subsection{Method Overview}
\label{sec:mo}
The LiDAR sensor of robotic systems collects a sequence of 3D point clouds $\mathcal{P}$. For the $t$-th frame, 3D point cloud $\textbf{P}^{t}$ is represented with a coordinate set $\textbf{C}^{t}$ and a feature set $\textbf{F}^{t}$. 
Note that $\textbf{F}^{t}$ is typically a set of RGB features (RGB-D camera), remission features (LiDAR sensor) or high dimension features processed by some learning modules. 
Then, we predict the labels for each point $\textbf{p}^{t}_{i}$. Here $\textbf{p}^{t}_{i}$ represents the $i$-th point at $\textbf{c}^{t}_{i}$, $\textbf{c}^{t}_{i} \in \textbf{C}^{t}$. $\textbf{p}^{t}_{i}$ contains two parts, the coordinate $\textbf{c}^{t}_{i}$ and the feature $\textbf{f}^{t}_{i}$. 

To tackle the 4D point cloud segmentation task, SpSequenceNet designs a two-branch sparse U-Net to extract the features of $\textbf{P}^{t}$ and $\textbf{P}^{t-1}$. 
Then, SpSequenceNet applies two cross-frame modules, which are cross-frame global attention and cross-frame local interpolation, to extract temporal information. 
Cross-frame global attention gathers global information of $\textbf{P}^{t-1}$ to generate a channel-wise attention mask. 
The attention mask guides the $\textbf{P}^{t}$ branch to generate high-quality features of $\textbf{P}^{t}$.
Cross-frame local interpolation searches the nearest neighbours of the point $\textbf{p}^{t}_{i}$ in the previous frames $\textbf{P}^{t-1}$. Then, cross-frame local interpolation aggregates all the nearest neighbours with a weighted sum based on an approximate Euclid distance. The interpolated features are related to the local context information of $\textbf{c}^{t}_{i}$ in $\textbf{P}^{t-1}$. Based on the spatial coherence, cross-frame local interpolation fuses the interpolated features and the features of the target point with a simple multiple layer perceptron module. 
Afterwards, the decoder predicts the labels of $\textbf{P}^{t}$ with the features from previous modules. 

However, the performance of SpSequenceNet is limited by the simple spatial-temporal model and the backbone network. 
The modules proposed in SpSequenceNet are unable to capture the variation features, especially for the high-speed objects.
In the experiment of SpSequenceNet, the local temporal features benefit the environment objects and low speed objects, like buildings and moving-person, and the improvement of the moving objects is below expectation. 
Furthermore, the backbone network of SpSequenceNet is a simplified 7-layer submanifold sparse convolution U-Net~\cite{graham2017submanifold}. 
According to the results reported by SpSequenceNet, the feature quality of the backbone network is lower than most of the novel backbone networks, like KPConv~\cite{thomas2019kpconv}.

To improve the performance of SpSequenceNet, we design a temporal variation-aware interpolation module to replace the cross-frame local interpolation for a better network performance on the moving objects.
In the temporal variation-aware interpolation module, we explore the usage of temporal variation information in 4D point cloud segmentation.
The experimental results in Table~\ref{table:res_com_d} show the temporal variation information improves the segmentation quality for the moving objects.
Then, we design a lightweight structure, temporal voxel-point refiner, to capture the temporal variation on the point level.
The temporal voxel-point refiner builds a directed graph between two corresponding frames, and uses a graph convolution to propagate the label information from the previous frame to the current frame.
The directed graph associates the label information of the objects in corresponding frames to model the spatial coherence and temporal variation.
Moreover, voxelization transforms the continuous point coordinates into the discrete space while the transformation drops the geometric details of continuous point coordinates in every voxels.
The temporal voxel-point refiner rebuilds the details with point level information of previous frames.
Ultimately, we adapt a MinkowskiUNet as the backbone of our work.
MinkowskiUNet has shown promising results in many works. 
Recently, they updated a more efficient version with a GPU-based hash table, making MinkowskiUNet capable of large-scale tasks. Motivated by these findings, we re-implement the backbone network with MinkowskiUNet.

\begin{figure*}
    \centering
	\includegraphics[width=0.90\linewidth]{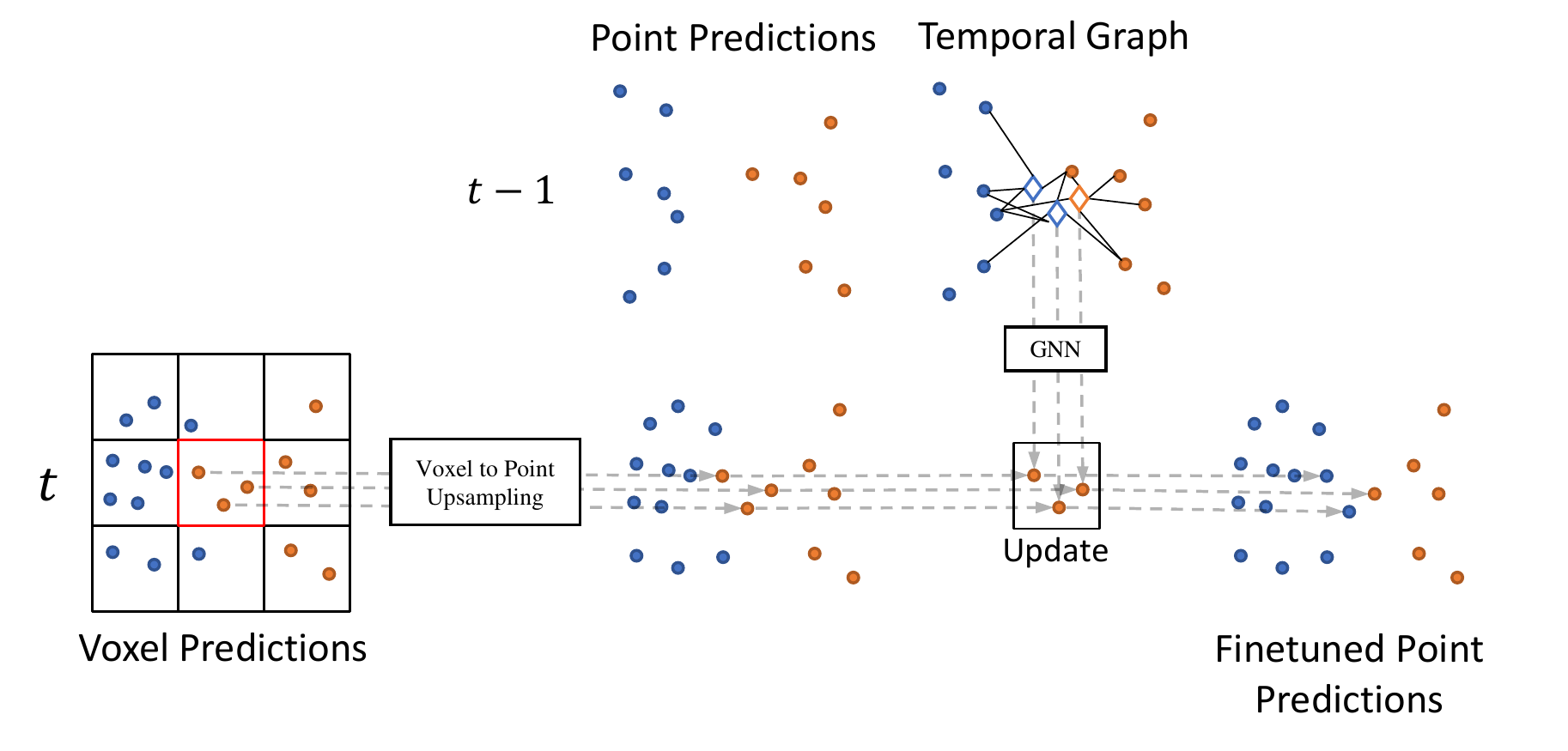}  

	\caption{\textbf{A simple example of the temporal voxel-point refiner.} Blue points and yellow points belong to two classes. $\diamondsuit$ indicates the coordinates of target points in Frame $t-1$. The temporal voxel-point refiner build a temporal graph between the point clouds $\textbf{P}^{t}$ and $\textbf{P}^{t-1}$. Then, based on the temporal graph, we adjust the voxel predictions to the point predictions. 
	}
	\label{fig:TP2V}

\end{figure*}

\begin{figure}
    \centering
	\includegraphics[width=0.6\linewidth]{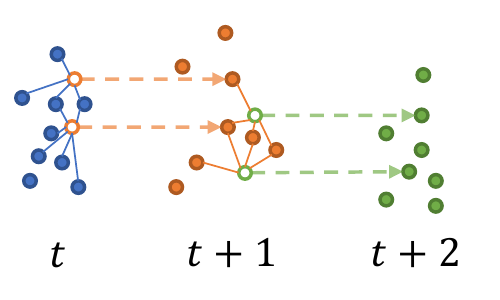}  

	\caption{\textbf{An example of temporal graph.} 
	}
	\label{fig:TP2V_points}

\end{figure}

\subsection{Temporal Variation-aware Interpolation}

Cross-frame local interpolation in SpSequenceNet generates an interpolated feature to represent the local information of the previous point cloud frame in the exact location of the current point cloud frame.
Then, cross-frame local interpolation concatenates the interpolated features and the features in the current point cloud frame as the input of the next module. 
However, the experimental results of cross-frame local interpolation show the improvement on moving objects is lower than expected. 
In our observation, the local features change drastically when the objects move fast.
The dynamic feature variation information is valuable for class recognition, especially for moving objects.
To this end, we modify the cross-frame local interpolation module to capture more temporal variation information between two corresponding point cloud frames. 
We show a comparison between the cross-frame local interpolation (SpSequenceNet) and the temporal variation-aware interpolation in Figure~\ref{fig:com_TVCLI}.

\noindent\textbf{Neighbor Selection and Distance Weights}. Following the SpSequenceNet, we search $k$ nearest neighbours of the target point $p^{t}_{i}$ in the previous point cloud frame with the same approximate Euclid distance in SpSequenceNet. 
\begin{equation}
    \textbf{D}^{t-1,t} = \frac {\textbf{C}^{t}\cdot (\textbf{C}^{t})^{T} + \textbf{C}^{t-1}\cdot (\textbf{C}^{t-1})^{T} - 2 \odot \textbf{C}^{t}\cdot (\textbf{C}^{t-1})^{T}} {\gamma^{2}}. 
\end{equation}
Here, $\odot$ is an element-wise multiplication.
$\textbf{D}^{t-1,t}$ is an approximate Euclid distance for the coordinate set $\textbf{C}^{t}$ and $\textbf{C}^{t-1}$. 
As the temporal variation-aware interpolation generates high level temporal features, the coordinate set $\textbf{C}^{t}$ and $\textbf{C}^{t-1}$ are voxelized to represent the position of points in the sparse tensor.
As the dimension of our input tensor is close to $128\times128\times128$, the values of each dimension in $\textbf{C}^{t}$ and $\textbf{C}^{t-1}$ belong to $[0,128]$.
The hyperparameter $\gamma$ of the approximate Euclid distance is set as 128 to normalized the distance matrix.
Based on $\textbf{D}^{t-1,t}$, we search the $k$ nearest neighbour set $\hat{\textbf{P}}^{t-1}_{i}$ of the target point $\textbf{p}^{t}_{i}$.
$\hat{\textbf{P}}^{t-1}_{i}$ also contains two parts, coordinates $\hat{\textbf{C}}^{t-1}_{i}$ and features $\hat{\textbf{F}}^{t-1}_{i}$.
Then, a interpolation weight of each point in $\hat{\textbf{P}}^{t-1}_{i}$ is formulated as:
\begin{equation}
    w_{j,i}^{t-1} = (\alpha  - \min(d^{t-1,t}_{j,i},\alpha)) * \beta,
\end{equation}
where $j$ is the index of nearest neighbour point in the previous frame.
$d^{t-1,t}_{j,i}$ is the approximate Euclid distance between $\textbf{c}_{i}^{t}$ and $\hat{\textbf{c}}_{j,i}^{t-1}$, $\hat{\textbf{c}}_{j,i}^{t-1} \in \hat{\textbf{C}}^{t-1}_{i}$.
$\alpha$ and $\beta$ are hpyer parameters. For $\alpha$ and $\beta$, we follow the original setting in SpSequenceNet, which are 0.5 and 2. 

\begin{figure*}
    \centering
	\includegraphics[width=0.65\linewidth]{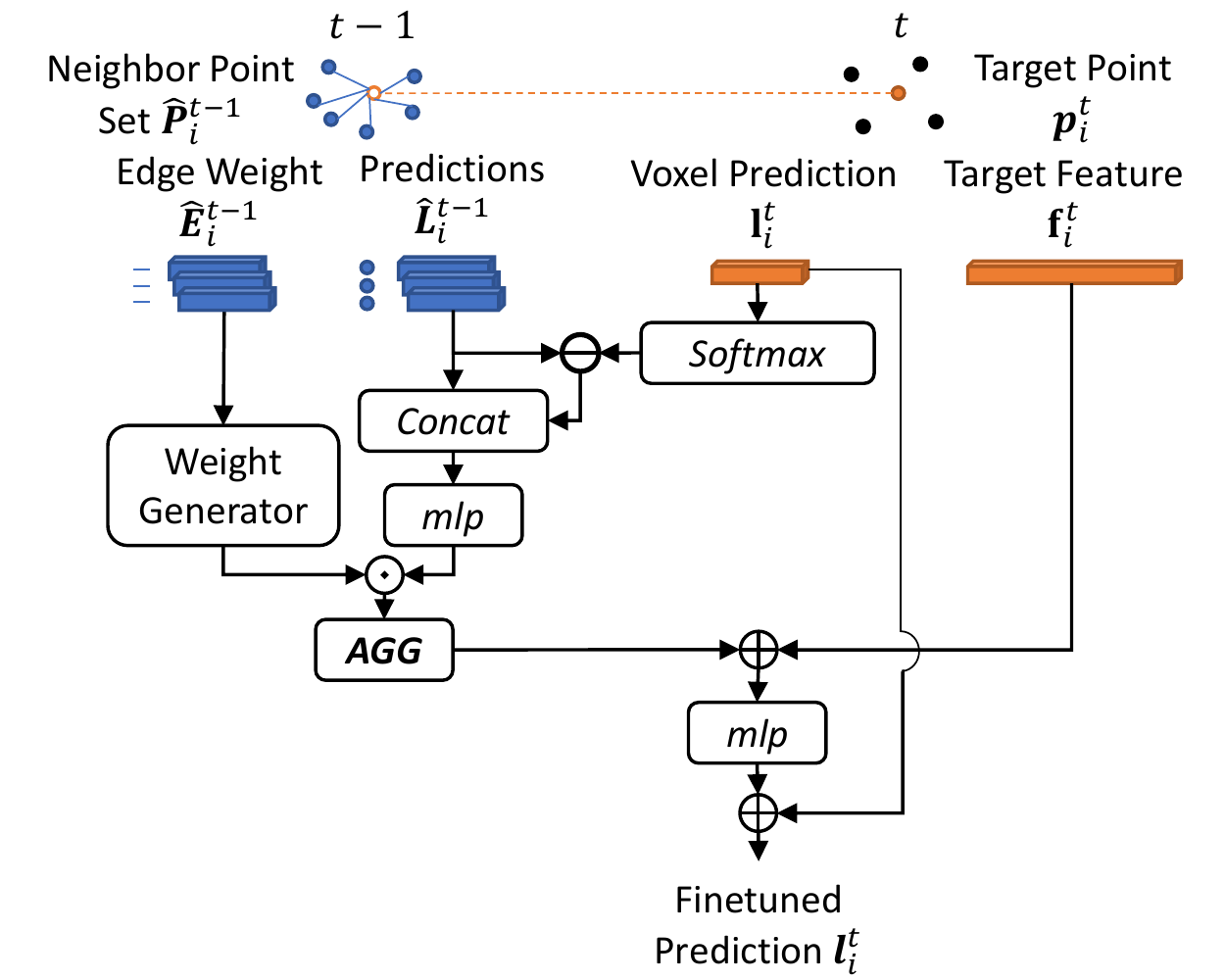}  

	\caption{\textbf{The structure of the temporal voxel-point refiner.} 
	$\hat{\textbf{P}}^{t-1}_{i}$ is the nearest neighbor set of the target point $\textbf{p}_{i}^{t}$ in Frame $t-1$. $\hat{\textbf{E}}^{t-1}_{i}$ is a set of the relative coordinate offset.
	$\hat{\textbf{L}}^{t-1}_{i}$ is the predictions of the nearest neighbours. 
	$\textbf{f}_{i}^{t}$ is the feature from the last decoder layer (before the prediction layer) and $\textbf{l}_{i}^{t}$ is the prediction of the target point. 
	$\odot$ is the element-wise multiplication. $\oplus$ is the element-wise plus, and $\ominus$ is the element-wise minus. $\textbf{\textit{AGG}}$ is an aggregation function, which is a maximum function in our module.
	}
	\label{fig:TP2V_main}
\end{figure*}

\noindent\textbf{Temporal Variation-aware interpolation}. With the nearest neighbour set and the interpolation weight, we use the features of previous frames and the feature $\textbf{f}_{i}^{t}$ of target point to generate the interpolation points.
Compared to our version, the cross-frame local interpolation only use the features of previous frames.
As stated in Section~\ref{sec:introduction}, the temporal variation is related to the difference of local regions in different frame.
We calculate the difference of the feature $\textbf{f}_{i}^{t}$ of target point and the features $\hat{\textbf{f}}_{j,i}^{t-1}$ of nearest neighbors to represent the temporal variation. Then, the variation-aware features $\hat{\textbf{v}}_{j,i}^{t-1}$ are formulated as following:
\begin{equation}
    \begin{split}
       \hat{\textbf{v}}_{j,i}^{t-1} =
       ReLU(mlp([\hat{\textbf{f}}_{j,i}^{t-1},       \textbf{f}_{i}^{t}-\hat{\textbf{f}}_{j,i}^{t-1}])).
    \end{split}
    \label{eq:i3}
\end{equation}
Here we concatenate $\hat{\textbf{f}}_{j,i}^{t-1}$ and $\textbf{f}_{i}^{t}-\hat{\textbf{f}}_{j,i}^{t-1}$ as the input of the multi-layer perceptron.
Afterward, we aggregate all the features $\hat{\textbf{f}}_{j,i}^{t-1}$ to generate the interpolated feature $\textbf{h}_{i}^{t}$, which is formulated as:
\begin{equation}
\textbf{h}_{i}^{t} = \sum_{j}^{k} w_{j,i}^{t-1}  \odot  \hat{\textbf{v}}_{j,i}^{t-1}.
\end{equation}
$\odot$ is an element-wise multiplication, and $w_{j,i}^{t-1}$ is the distance weights from previous step. 
$k$ is the number of nearest neighbours from the previous step.
Then, we concatenate the interpolated feature $\textbf{h}_{\textbf{c}_{t,i}}$ and $\textbf{f}_{i}^{t}$ together as the input of the decoder.

\subsection{Temporal Voxel-Point Refiner}

Recent works~\cite{xu2021rpvnet,cheng20212} in 3D point cloud segmentation shows the point level details are valuable for the segmentation task.
Inspired by these observations, we dig in the usage of point level details to capture the spatial coherence and temporal variation.
As our backbone network is a volumetric network, voxelization discretizes the point coordinates and drops the point level details.
To further make use of point level details, we design a lightweight graph neural network to extract spatial and temporal information. 
We build a directed temporal graph on two corresponding point cloud frames.
For each point in the current frame, we search k nearest neighbours in the previous frame.
The directed edges of the directed temporal graph are from k nearest neighbours in the previous frame to the point in the current frame.
Then, a temporal graph convolution is designed to propagate the information of each node to the next frame.
For each propagation step, we refine the predictions of current frame for better segmentation quality. We also show an explanation in Figure~\ref{fig:TP2V}.

\noindent\textbf{Temporal Graph Building}. We build an oriented graph on the previous and current frames to model spatial and temporal information. For each point $\textbf{p}_{i}^{t}$, we compute a $k$ nearest neighbour sub-graph in the previous frame. In the k nearest neighbour sub-graph, the nodes are the nearest neighbours $\hat{\textbf{P}}^{t-1}_{i}$ and the target point $\textbf{p}_{i}^{t}$.
The edges are directed connections from the nearest neighbours to the target point $\textbf{p}_{i}^{t}$. 
We show a visualization of the temporal graph in Figure~\ref{fig:TP2V_main}. 
The oriented temporal graph propagates the spatial and temporal information along the sequence.
The extraction of motion details in each small region relies on the the spatial and temporal information from previous frames.
Furthermore, the point level sub-graph contains the geometric details of the points from previous frame. 
The local geometric details improves the performance of network on small objects.

\noindent\textbf{Temporal Graph Convolution}.
After building the temporal graph, we design a temporal graph convolution to update the predictions of target point $\textbf{p}_{i}^{t}$ with the information from previous frame. 
We show the structure of the temporal voxel-point refiner in Figure~\ref{fig:TP2V_main}.
For each sub-graph, we compute edge weights $\textbf{e}_{j}^{t-1}$ for the nearest neighbours $\hat{\textbf{P}}_{i}^{t-1}$. 
Firstly, a relative position weight $\textbf{d}_{j}^{t-1}$ is computed as following:
\begin{equation}
    \begin{array}{r c l}
        \Delta\textbf{c}_{j,i}^{t-1,t} &=& \tanh(\hat{\textbf{c}}_{j}^{t-1}-\textbf{c}_{i}^{t}),\\ 
        \textbf{d}_{j}^{t-1} &=& 1-| \Delta\textbf{c}_{j,i}^{t-1,t}|.
  \end{array}
\end{equation}
Here, $\textbf{d}_{j}^{t-1}$ is a three-dimension vector, representing the relative position weights for $x$, $y$, and $z$ dimensions. 
The relative position weights $\textbf{d}_{j}^{t-1}$  belong to $(0,1]$. 
With this setting, the relative position weight $\textbf{d}_{j}^{t-1}$ helps to encode a location-sensitive edge weight for the temporal graph convolution.
Then, we generate an edge weight $\textbf{e}_{j}^{t-1}$, which is formulated as
\begin{equation}
    \textbf{e}_{j}^{t-1} = \tanh{(mlp(\textbf{d}_{j}^{t-1}))},
\end{equation}
where the output dimension of $mlp$ function is the same as the dimension of the target point features $\textbf{f}_{i}^{t}$.
Then, we aggregate all the features of nodes together to update the network output $\textbf{l}_{i}^{t}$ of the target point. In this aggregation step, we use the predictions $\textbf{y}_{j}^{t-1}$ as the features of neighbours.
Another option is to use the output features $\textbf{f}_{j}^{t-1}$ before the final prediction layer as the the features of neighbours. 
With comparison between two options, our current implementation requires less memory and computation costs to martian and update the temporal graph.
Afterwards, we generate the adjusted features of nodes with the edge weight $\textbf{e}_{j}^{t-1}$, the target feature $\textbf{y}_{i}^{t}$ and the features of nodes.
To aggregate the adjusted features of nodes, we define the temporal graph convolution by applying a channel-wise maximum function on the adjusted features. 
In particular, the updated features $\textbf{f}'_{\hat{\textbf{c}}_{i}^{t}}$ are:
\begin{equation}
    \mathbf{\mathit{f}}_{i}^{t} = \textbf{f}_{i}^{t}+\mathcal{M}(\textbf{e}_{j}^{t-1}\odot mlp([\textbf{y}_{j}^{t-1},\textbf{y}_{i}^{t}-\textbf{y}_{j}^{t-1}])).
\end{equation}
$\mathcal{M}$ is a channel-wise maximum function, $\odot$ is an element-wise multiplication, and $\textbf{y}_{i}^{t}$ is the class probability of the target point $\textbf{p}_{i}^{t}$ from the network. 
Then, with the updated features $\mathbf{\mathit{f}}_{i}^{t}$, we finetune the final outputs:
\begin{equation}
    \mathbf{\mathit{l}}_{i}^{t} = \textbf{l}_{i}^{t} + mlp(\mathbf{\mathit{f}}_{i}^{t}),
\end{equation}
where $\textbf{l}_{i}^{t}$ is the raw predictions without softmax layer from the sparse convolution U-Net.

Our temporal voxel-point refiner utilizes the temporal information and the point coordinates efficiently. 
Compared to other methods described in Section~\ref{sec:introduction}, our temporal voxel-point refiner avoids using an encoder-decoder structure for fusing point and voxel features, and only requires 33k parameters to achieve a significant improvement.
Furthermore, directly setting the features of neighbours $\hat{\textbf{P}}_{i}^{t}$ with the predictions $\textbf{L}_{\hat{\textbf{c}}^{t-1}_{j}}$ requires low memory and computation costs. In our implementation, the averaging inference time of the temporal voxel-point refiner is around $3ms$ when the amount of input data is around $200k$ points.

\subsection{Two-Stage Training}
As the input of the temporal voxel-point refiner requires the predictions from previous frames, the traditional batch training shows some disadvantages. We design a two-stage training pipeline to train our network to obtain a robust model.

\textbf{Single-frame Pre-training}. Similar to image-based methods, an excellent pre-trained model affects the performance and stabilizes the training of the main framework. 
We pre-train our backbone network as a single frame backbone network on the multiple frame task of SemanticKITTI. 
Although several labels are related to motion status in multiple frame tasks, the backbone network can still capture some motion status.
In our practice, the pre-trained model reduces the training time in the next step. 

\textbf{Multi-frame Training}. We follow the SpSequenceNet to train our two-branch network. Initially, we align the current frame and previous frame in the same coordinate system with the pose information.
Then, a random rotation with the same random seed is applied to the current and previous frames.
Unlike SpSequenceNet, we also apply a random point dropping to accelerate the training process in the multi-frame training stage. The random dropping reduces the computational resources and does not bring a performance degradation. 
Afterwards, we train our network with a standard training framework.

\textbf{Temporal Voxel-point Refiner Training}. The temporal voxel-point pipeline requires the predictions from the previous frame. However, in our practice, the overfitting problem is caused by directly training the temporal voxel-point refiner with the ground truth from the previous frame. 
Furthermore, it requires a massive computation racecourse to simultaneously train the temporal voxel-point refiner and the main network.
Therefore, we only train the temporal voxel-point refiner in the current stage.
Initially, we load the best model from the previous stage and freeze all the parameters of the main network. 
Then, during the training phase, we use the ground truth as the input when it is the first time for this frame to be loaded. 
Afterwards, the predictions are the pseudo prediction for the next training step of this frame. Note that we use the first frame $\textbf{P}^{0}$ itself as the previous frame $\textbf{P}^{-1}$ of $\textbf{P}^{0}$, and the pseudo prediction of $\textbf{P}^{0}$ is the prediction of $\textbf{P}^{0}$ from the multi-frame training stage. The training of the temporal voxel-point refiner converges rapidly.

\textbf{Training detail}. 
As the point numbers of each class are unbalanced, a focal loss is used to restrain the effect of data unbalance. 
Then, we minimize the focal loss with an Adam optimizer. 
We use a standard desktop with a single Nvidia RTX 3090. Single-frame pre-training of a 34-layer backbone network takes about two days, and multi-frame training of the 34-layer model takes more than six days. Then, training a temporal voxel-point refiner requires about two days.

\begin{table*}
	\begin{center}
		\resizebox{\textwidth}{!}{%
			
			\begin{tabular}{l|c|ccccccccccccccccccccccccc}
				\hline
				& \rotatebox{90}{\textbf{mIoU}} & \rotatebox{90}{car} & \rotatebox{90}{bicycle} & \rotatebox{90}{motorcycle} & \rotatebox{90}{truck} & \rotatebox{90}{other-vehicle} & \rotatebox{90}{person} & \rotatebox{90}{bicyclist}&\rotatebox{90}{ motorcyclist} & \rotatebox{90}{road} & \rotatebox{90}{parking} & \rotatebox{90}{sidewalk} & \rotatebox{90}{other-ground} & \rotatebox{90}{building} & \rotatebox{90}{fence} & \rotatebox{90}{vegetation} & \rotatebox{90}{trunk} & \rotatebox{90}{terrain} & \rotatebox{90}{pole} & \rotatebox{90}{traffic-sign} & \rotatebox{90}{moving-car} & \rotatebox{90}{moving-bicyclist} & \rotatebox{90}{moving-person} & \rotatebox{90}{moving-motorcyclist} & \rotatebox{90}{moving-other-vehicle} & \rotatebox{90}{moving-truck}  \\
				\hline\hline
				TangentConv~\cite{tatarchenko2018tangent} & 34.1 & 84.9 & 2.0 & 18.2 & 21.1 & 18.5 & 1.6 & 0.0 & 0.0 & 83.9 & 38.3 & 64.0 & 15.3 & 85.8 & 49.1 & 79.5 & 43.2 & 56.7 & 36.4 & 31.2 & 40.3 & 1.1 & 6.4 & 1.9 & \textbf{30.1} & \textbf{42.2}  \\
				
				DarkNet53Seg~\cite{behley2019semantickitti} & 41.6 & 84.1 & 30.4 & 32.9 & 20.0 & 20.7 & 7.5 & 0.0 & 0.0 & 91.6 & 64.9 & 75.3 & \textbf{27.5} & 85.2 & 56.5 & 78.4 & 50.7 & 64.8 & 38.1 & 53.3 & 61.5 & 14.1 & 15.2 & 0.2 & 28.9 & 37.8   \\
				TemporalLidarSeg~\cite{duerr2020lidar} & 47.0 & 92.1 & 47.7 & 40.9 & \textbf{39.2} & 35.0 & 14.4 & 0.0 & 0.0 & \textbf{91.8} & 59.6 & \textbf{75.8} & 23.2 & 89.8 & 63.8 & 82.3 & 62.5 & 64.7 & 52.6 & 60.4 & 58.2 & 42.8 & 40.4 & 12.9 & 12.4 & 2.1  \\
				\hline
				SpSequenceNet~\cite{shi2020spsequencenet} & 43.1 & 88.5 & 24.0 & 26.2 & 29.2 & 22.7 & 6.3 & 0.0 & 0.0 & 90.1 & 57.6 & 73.9 & 27.1 & 91.2 & 66.8 & 84.0 & 66.0 & 65.7 & 50.8 & 48.7 & 53.2 & 41.2 & 26.2 & 36.2 & 2.3 & 0.1  \\
				TVSN(Ours) & \textbf{52.5} & \textbf{93.5} & 45.6 & \textbf{42.8} & 36.9 & \textbf{35.7} & \textbf{15.2} & 0.0 & 0.0 & 89.9 & \textbf{67.9} & 73.5 & 21.4 & \textbf{92.3} & \textbf{69.5} & \textbf{85.0} & \textbf{70.3} & \textbf{68.5} & \textbf{61.2} & \textbf{65.4} & \textbf{74.2} & \textbf{65.5} & \textbf{55.6} & \textbf{63.0} & 14.8 & 5.2  \\
				\hline
		\end{tabular}}
	\end{center}
	\caption{\textbf{Our test results on the multiple scan segmentation task of SemanticKITTI.} Note that SpSequenceNet is our backbone network.
	For the previous works, the performance of our proposed method is 5.5\% higher than TemporalLidarSeg~\cite{duerr2020lidar}. Compared to other method, our method shows an improvement in most of small or moving objects, like person, traffic sign, and moving car.}
	\label{table:res_total}
\end{table*} 

\begin{table}[t] 
	\begin{center}
		\resizebox{0.35\textwidth}{!}{%
			\begin{tabular}{l|c}
				\hline
				& mIoU\\
				\hline\hline
				PointNet++~\cite{qi2017pointnetplusplus}& 20.1\\
    			RandLA-Net~\cite{hu2020randla} & 53.5\\
    			KPConv~\cite{thomas2019kpconv} & 55.2\\
    			JS3C-Net~\cite{yan2020sparse} & 60.2\\
				\hline
				Backbone-34+TVI & 61.9 \\
				Backbone-34+TVI+TVPR & \textbf{63.0}\\
				\hline
		\end{tabular}}
	\end{center}
	\caption{\textbf{The results on the data part 3 of SemanticPOSS.} }
	\label{table:res_poss}
\end{table}

\begin{table}[t] 
	\begin{center}
		\resizebox{0.36\textwidth}{!}{%
			\begin{tabular}{l|c}
				\hline
				& mIoU\\
				\hline\hline
				MinkUNet-14 & 47.5\\
				MinkUNet-34 & 51.0\\
				SpSequenceNet& 43.1\\
				\hline

				Backbone-14 & 50.0\\
				Backbone-34 & 51.5\\
				Backbone-14+TVI & 50.7\\
				Backbone-34+TVI & 52.3\\
			    Backbone-14+TVI+TVPR & 52.4\\
				Backbone-34+TVI+TVPR & \textbf{53.8}\\
				\hline
		\end{tabular}}
	\end{center}
	\caption{\textbf{The ablation study of the validation dataset of SemanticKITTI.} We implement a new MinkUNet-based SpSequenceNet as the backbone network. Note that the performance of SpSequenceNet is directly cited from the original paper~\cite{shi2020spsequencenet}.}
	\label{table:res_com}
\end{table}

\section{Experiments}
\label{sec:test}
We evaluate our model on the multiple scan segmentation task on two outdoor sequence point cloud datasets, SemanticKITTI~\cite{behley2019semantickitti} and SemanticPOSS~\cite{pan2020semanticposs}. 
For SemanticKITTI, we follow the official split and evaluate our experiments on the test split (Section~\ref{sec:test_kitti}) and the validation split (Section~\ref{sec:validate_kitti}). 
For the point clouds in SemanticKITTI, the voxelization unit is $0.05 m$, and the size of one voxelized point cloud is around $3000 \times 3000 \times 1000$. 
For SemanticPOSS, we use data part 3 (500 frames) as the test dataset. SemanticPOSS contains a high amount of small objects. 
Furthermore, as the data collection device of SemanticPOSS is a 40-line LiDAR, the point cloud resolution is lower than the resolution in KITTI~\cite{geiger2012we} (128-line LiDAR). Therefore, small objects and lower resolution increase the challenge in segmentation. For the point clouds in SemanticPOSS, the voxelization unit is $0.1 m$. For better performance, we also finetune the main network trained with the model of SemanticKITTI.  

\subsection{Comparison}
\label{sec:test_kitti}
\textbf{The Multiple Scan Task of SmeanticKITTI}. We show results on the multiple scan segmentation test set of SemanticKITTI in Figure~\ref{table:res_total}. Our method clearly outperforms all existing methods in terms of mIoU, which is 9.4\% higher than SpSequenceNet and 5.5\% higher than TemporalLidarSeg. Compared to other 4D semantic segmentation networks, our method approach significantly improves performances on most small or moving objects. For example, our performances on moving people, bicyclists and motorcyclists are more than 10\% higher than other methods.

\label{sec:test_poss}
\noindent\textbf{SemanticPOSS}. Similar to the experiment of semanticKITTI, we apply a two-stage training on the semanticPOSS. We also report the training results for Backbone-34+TVI and Backbone-34+TVI+TVPR. Compared to other novel structures, the performance of our method is 2.8\% higher than JS3C-Net~\cite{yan2020sparse}. Without the temporal voxel-point refiner, Backbone-34+TVI achieves a 1.7\% improvement. Note that KPConv~\cite{thomas2019kpconv}, RandLA-Net~\cite{hu2020randla} and PointNet++~\cite{qi2017pointnetplusplus} are designed for single-frame scenarios and use one frame per iteration. JS3C-Net is designed to solve both semantic segmentation and scene completion. Furthermore, JS3C-Net align corresponding 70 frames and inference based on location in the aligned frames.

\subsection{Ablation Experiments}
\label{sec:validate_kitti}
\begin{figure*}
    \centering
	\includegraphics[width=0.8\linewidth]{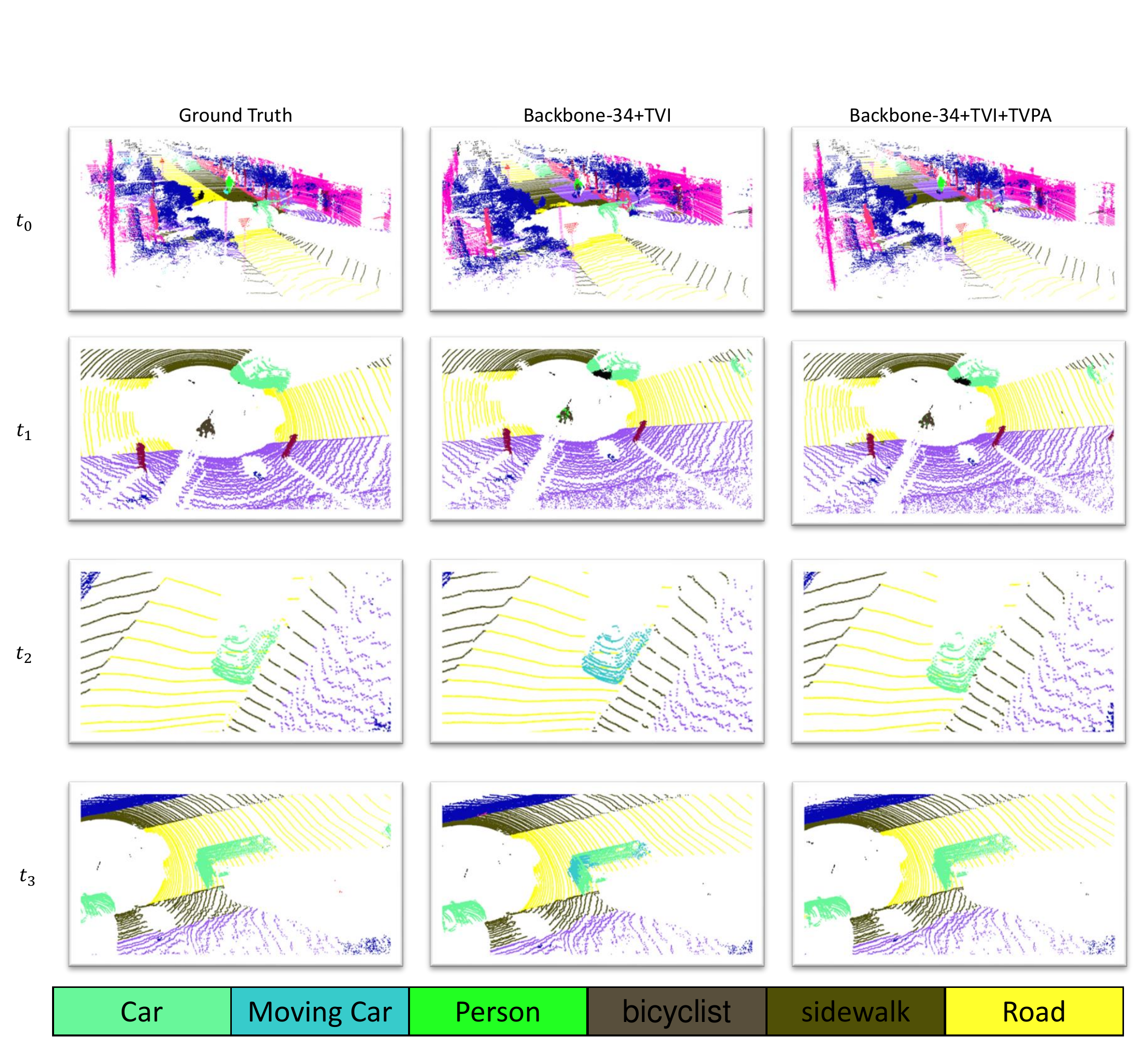}  

	\caption{\textbf{Qualitative results for small object.} 
	}
	\label{fig:qc_s}

\end{figure*}

As shown in Table\ref{table:res_com}, we apply an ablation study on the validation dataset of SemanticKITTI. Then, we also analyze the effects of the temporal voxel-point refiner with some selected classes, which is demonstrated in Figure~\ref{fig:qc_s}.
 
\textbf{The backbone network}. We adapt MinkowskiUNet to implement a new version of SpSequenceNet as the backbone network of our method. 
The performance of SpSequenceNet with MinkowskiUNet-34 (backbone 34) is 7.6\% higher than SpSequenceNet and 0.5\% higher than MinkowskiUNet-34. Note that MinkowskiUNet-34 is a 34 layer 3D UNet.
Furthermore, the performance of Backbone-14 is 6.9\% higher than SpSequenceNet, and 2.5\% higher than MinkowskiUNet-14. 

\textbf{Backbone+TVI}. Here we modify the cross-frame local interpolation with our proposed temporal variation-aware interpolation in the backbone network. The temporal variation-aware interpolation achieves a 0.7\% improvement for backbone-14 and a 0.8\% improvement for backbone-34.
In our experiment, we try two versions of TVI with 5 neighbours and 7 neighbours. 
However, similar to the report in SpSequenceNet, more neighbours do not significantly benefit the performance of cross-frame local interpolation. 
To balance the computation and the precision, we use 5 as the number of nearest neighbours in this experiment.

\textbf{Backbone+TVI+TVPR}. The temporal voxel-point refiner has a minimum 1.5\% improvement. 
The temporal voxel-point refiner extracts the details of points and captures more temporal information on the point level. 
Similar information is not considered in backbone+TVI. 
Therefore, the temporal voxel-point refiner shows a significant improvement.

\textbf{Analysis of temporal variation modules}. As stated in Table~\ref{table:res_com_d}, we select six classes related to the undersized and moving objects. Each selected class owns more than $10k$ points in the validation set. Then, we evaluate the IoU for each class. 
For TVI, although there are drawbacks for people and bicycles, the temporal variation benefits all moving classes and the overall performance of TVI is 0.8\% higher than the backbone network. 
Then, TVPR shows a noticeable improvement on moving objects with the point level temporal information.
On the other hand, our TVPR rebuilds the point level details.
Compared to the results of Backbone-34+TVI, TVPR also has higher performance on small objects, e.g. person and bicycle.

\begin{figure*}
    \centering
	\includegraphics[width=0.95\linewidth]{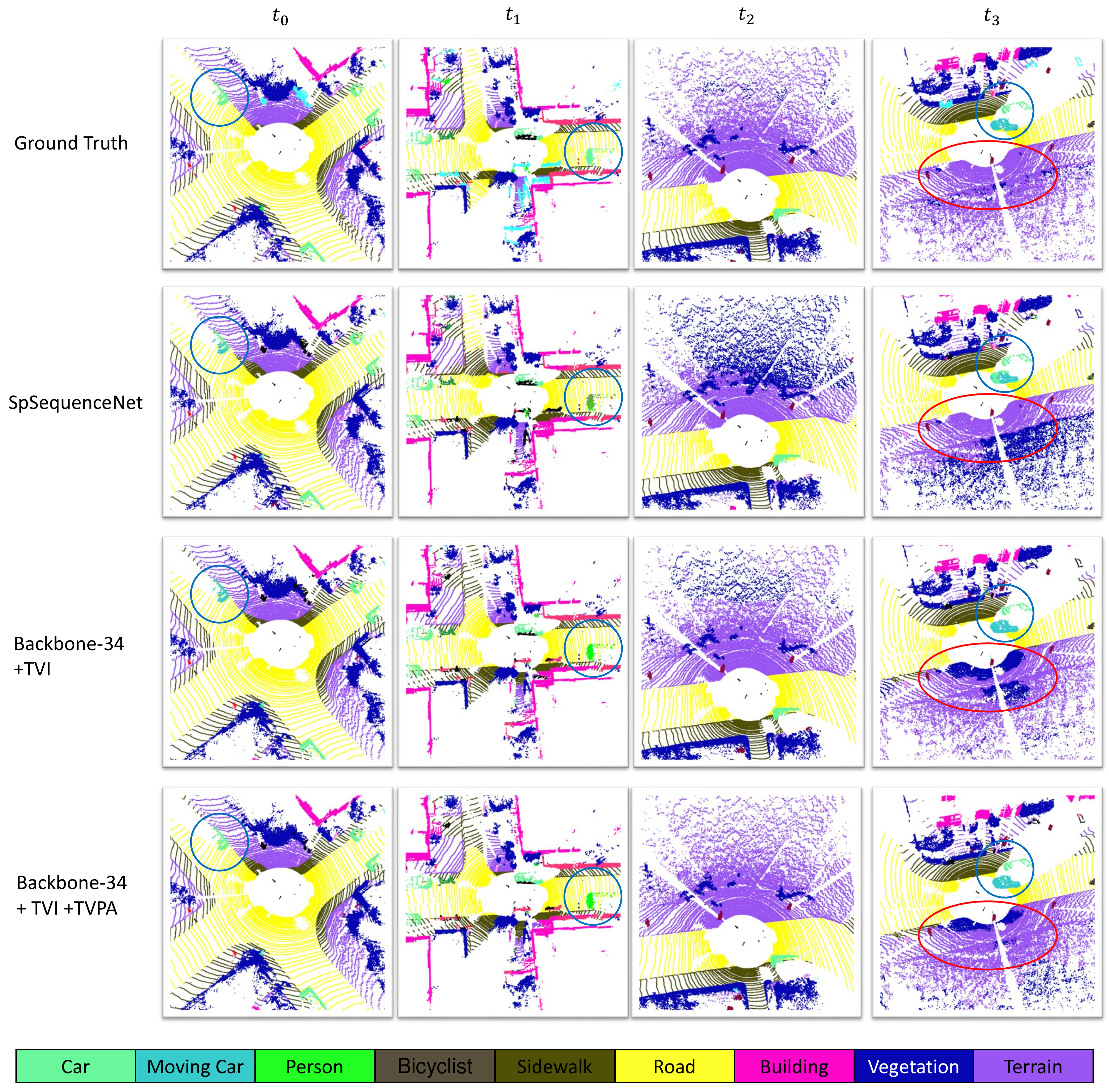}  

	\caption{\textbf{Qualitative results on SemanticKITTI.} In the blue circle, our prediction of small objects is better than SpSequenceNet. There is a failure case for all three method in the red circle.
	}
	\label{fig:qc_main}

\end{figure*}

\begin{table*}[t] 
	\begin{center}
		\resizebox{1.\textwidth}{!}{%
			\begin{tabular}{l|c|cccccccc}
				\hline
				& mIoU & car & person & bicycle  & moving-car & moving-person & moving-bicyclist\\
				\hline\hline
				Backbone-34 & 51.5 & 93.8 & 24.1 & 29.4  & 71.4 & 54.8 & 90.2\\
				Backbone-34+TVI & 52.3 & 94.4 & 18.1 & 28.0 & 73.6 & 55.7 & 90.7\\
				Backbone-34+TVI+TVPR & 53.8 & 94.2 & 20.0 & 40.4  & 78.3 & 57.1 & 94.7\\
				\hline
		\end{tabular}}
	\end{center}
	\caption{\textbf{An analysis for temporal variation modules.}  
	}
	\label{table:res_com_d}
\end{table*}

\subsection{Qualitative Results}
Figure~\ref{fig:qc_main} shows some visualized results from SemanticKITTI. 
We compare our proposed method with SpSequenceNet. In frame $t_{1}$, $t_{2}$ and $t_{3}$, SpSequenceNet fails to segment the environment classes. 
Furthermore, in frame $t_{0}$, frame $t_{2}$ and frame $t_{3}$, some small objects (\textit{i.e., moving car}) are classified as multiple other classes by SpSequenceNet. 
By comparison, our Backbone-34+TVI shows the capability to capture high-quality features. 
Furthermore, in frame $t_{3}$, TVI captures the moving status of objects and achieves a better performance.
Then, the temporal voxel-point refiner improves the smoothness of the prediction and shows a higher capability to capture the moving and non-moving objects. In frame $t_{0}$, the temporal voxel-point refiner adjusts the prediction of the car in the blue circle as a parked car while other methods segment the same area with both the parked car and moving car. 
The quantitative and qualitative results clearly verify the effectiveness of the proposed methods.

Besides, we also demonstrate several examples of small objects in Figure~\ref{fig:qc_s}. 
In frame $t_0$ and frame $t_1$, Backbone-34+TVI predicts inconsistent results for the same objects. The inconsistent result is a common problem in the volumetric-based method, attributed to the detail loss. 
During the voxelization step, the details of small objects are missing and increase the difficulties of feature extraction for the volumetric-based method. 
As a result, our temporal voxel-point refiner efficiently improves the performance of the volumetric-based methods, which can be verified in the result of Backbone-34+TVI+TVPR.
In frame $t_2$ and frame $t_3$, Backbone-34+TVI also fails to segment the motion status of the objects. In our observation, the deviation from the voxelization, the alignment of two frames and the downsampling during the sparse network blurs the motion information between two corresponding frames. Therefore, the deviation limits the ability for cross-frame local interpolation and temporal variation-aware interpolation. Consequently, our temporal voxel-point refiner propagates the motion status at a point-to-point level, reducing the deviation effects. We can verify the effectiveness of the temporal voxel-point refiner with a better performance about the moving classes in frame $t_2$ and $t_3$.

\section{Conclusion}
In this paper, we have presented a spatial-temporal network to improve the 4D perception ability. 
To further dig into the usage of temporal information, We propose a temporal variation-aware interpolation module to capture more variation information between corresponding frames. 
Furthermore, we design a temporal voxel-point refiner with a temporal graph structure to capture the temporal variation and spatial coherence. The temporal voxel-point refiner can also reduce the error from the voxelization. %
The experiments have shown that each component of our method is highly effective and we achieve state-of-the-art results on SemanticKITTI and SemanticPOSS.

\ifCLASSOPTIONcompsoc
  \section*{Acknowledgments}
\else
  \section*{Acknowledgment}
\fi
This research is supported by the National Research Foundation, Singapore under its AI Singapore Programme (AISG Award No: AISG-RP-2018-003), and the Ministry of Education, Singapore, Tier-1 research grants: RG28/18 (S), RG22/19 (S) and RG95/20.

\ifCLASSOPTIONcaptionsoff
  \newpage
\fi

{\small
\bibliographystyle{IEEEtran}
\bibliography{bare_jrnl_compsoc}

\begin{thebibliography}{10}
\providecommand{\url}[1]{#1}
\csname url@samestyle\endcsname
\providecommand{\newblock}{\relax}
\providecommand{\bibinfo}[2]{#2}
\providecommand{\BIBentrySTDinterwordspacing}{\spaceskip=0pt\relax}
\providecommand{\BIBentryALTinterwordstretchfactor}{4}
\providecommand{\BIBentryALTinterwordspacing}{\spaceskip=\fontdimen2\font plus
\BIBentryALTinterwordstretchfactor\fontdimen3\font minus
  \fontdimen4\font\relax}
\providecommand{\BIBforeignlanguage}[2]{{%
\expandafter\ifx\csname l@#1\endcsname\relax
\typeout{** WARNING: IEEEtran.bst: No hyphenation pattern has been}%
\typeout{** loaded for the language `#1'. Using the pattern for}%
\typeout{** the default language instead.}%
\else
\language=\csname l@#1\endcsname
\fi
#2}}
\providecommand{\BIBdecl}{\relax}
\BIBdecl

\bibitem{xu2021rpvnet}
J.~Xu, R.~Zhang, J.~Dou, Y.~Zhu, J.~Sun, and S.~Pu, ``Rpvnet: A deep and
  efficient range-point-voxel fusion network for lidar point cloud
  segmentation,'' \emph{arXiv preprint arXiv:2103.12978}, 2021.

\bibitem{cheng20212}
R.~Cheng, R.~Razani, E.~Taghavi, E.~Li, and B.~Liu, ``2-s3net: Attentive
  feature fusion with adaptive feature selection for sparse semantic
  segmentation network,'' in \emph{Proceedings of the IEEE/CVF Conference on
  Computer Vision and Pattern Recognition}, 2021, pp. 12\,547--12\,556.

\bibitem{shi2020spsequencenet}
H.~Shi, G.~Lin, H.~Wang, T.-Y. Hung, and Z.~Wang, ``Spsequencenet: Semantic
  segmentation network on 4d point clouds,'' in \emph{Proceedings of the
  IEEE/CVF Conference on Computer Vision and Pattern Recognition}, 2020, pp.
  4574--4583.

\bibitem{duerr2020lidar}
F.~Duerr, M.~Pfaller, H.~Weigel, and J.~Beyerer, ``Lidar-based recurrent 3d
  semantic segmentation with temporal memory alignment,'' in \emph{2020
  International Conference on 3D Vision (3DV)}.\hskip 1em plus 0.5em minus
  0.4em\relax IEEE, 2020, pp. 781--790.

\bibitem{puy2020flot}
G.~Puy, A.~Boulch, and R.~Marlet, ``Flot: Scene flow on point clouds guided by
  optimal transport,'' in \emph{Computer Vision--ECCV 2020: 16th European
  Conference, Glasgow, UK, August 23--28, 2020, Proceedings, Part XXVIII
  16}.\hskip 1em plus 0.5em minus 0.4em\relax Springer, 2020, pp. 527--544.

\bibitem{liu2019meteornet}
X.~Liu, M.~Yan, and J.~Bohg, ``Meteornet: Deep learning on dynamic 3d point
  cloud sequences,'' in \emph{Proceedings of the IEEE/CVF International
  Conference on Computer Vision}, 2019, pp. 9246--9255.

\bibitem{behley2019semantickitti}
J.~Behley, M.~Garbade, A.~Milioto, J.~Quenzel, S.~Behnke, C.~Stachniss, and
  J.~Gall, ``Semantickitti: A dataset for semantic scene understanding of lidar
  sequences,'' in \emph{Proceedings of the IEEE International Conference on
  Computer Vision}, 2019, pp. 9297--9307.

\bibitem{pan2020semanticposs}
Y.~Pan, B.~Gao, J.~Mei, S.~Geng, C.~Li, and H.~Zhao, ``Semanticposs: A point
  cloud dataset with large quantity of dynamic instances,'' in \emph{2020 IEEE
  Intelligent Vehicles Symposium (IV)}.\hskip 1em plus 0.5em minus 0.4em\relax
  IEEE, 2020, pp. 687--693.

\bibitem{lawin2017deep}
F.~J. Lawin, M.~Danelljan, P.~Tosteberg, G.~Bhat, F.~S. Khan, and M.~Felsberg,
  ``Deep projective 3d semantic segmentation,'' in \emph{International
  Conference on Computer Analysis of Images and Patterns}.\hskip 1em plus 0.5em
  minus 0.4em\relax Springer, 2017, pp. 95--107.

\bibitem{boulch2017unstructured}
A.~Boulch, B.~Le~Saux, and N.~Audebert, ``Unstructured point cloud semantic
  labeling using deep segmentation networks.'' \emph{3DOR}, vol.~2, p.~7, 2017.

\bibitem{wu2018squeezeseg}
B.~Wu, A.~Wan, X.~Yue, and K.~Keutzer, ``Squeezeseg: Convolutional neural nets
  with recurrent crf for real-time road-object segmentation from 3d lidar point
  cloud,'' in \emph{2018 IEEE International Conference on Robotics and
  Automation (ICRA)}.\hskip 1em plus 0.5em minus 0.4em\relax IEEE, 2018, pp.
  1887--1893.

\bibitem{wu2019squeezesegv2}
B.~Wu, X.~Zhou, S.~Zhao, X.~Yue, and K.~Keutzer, ``Squeezesegv2: Improved model
  structure and unsupervised domain adaptation for road-object segmentation
  from a lidar point cloud,'' in \emph{2019 International Conference on
  Robotics and Automation (ICRA)}.\hskip 1em plus 0.5em minus 0.4em\relax IEEE,
  2019, pp. 4376--4382.

\bibitem{xu2020squeezesegv3}
C.~Xu, B.~Wu, Z.~Wang, W.~Zhan, P.~Vajda, K.~Keutzer, and M.~Tomizuka,
  ``Squeezesegv3: Spatially-adaptive convolution for efficient point-cloud
  segmentation,'' \emph{arXiv preprint arXiv:2004.01803}, 2020.

\bibitem{cortinhal2020salsanext}
T.~Cortinhal, G.~Tzelepis, and E.~E. Aksoy, ``Salsanext: Fast,
  uncertainty-aware semantic segmentation of lidar point clouds for autonomous
  driving,'' 2020.

\bibitem{su2018splatnet}
H.~Su, V.~Jampani, D.~Sun, S.~Maji, E.~Kalogerakis, M.-H. Yang, and J.~Kautz,
  ``Splatnet: Sparse lattice networks for point cloud processing,'' in
  \emph{Proceedings of the IEEE Conference on Computer Vision and Pattern
  Recognition}, 2018, pp. 2530--2539.

\bibitem{graham2014spatially}
B.~Graham, ``Spatially-sparse convolutional neural networks,'' \emph{arXiv
  preprint arXiv:1409.6070}, 2014.

\bibitem{riegler2017octnet}
G.~Riegler, A.~Osman~Ulusoy, and A.~Geiger, ``Octnet: Learning deep 3d
  representations at high resolutions,'' in \emph{Proceedings of the IEEE
  Conference on Computer Vision and Pattern Recognition}, 2017, pp. 3577--3586.

\bibitem{graham2017submanifold}
B.~Graham and L.~van~der Maaten, ``Submanifold sparse convolutional networks,''
  \emph{arXiv preprint arXiv:1706.01307}, 2017.

\bibitem{choy20194d}
C.~Choy, J.~Gwak, and S.~Savarese, ``4d spatio-temporal convnets: Minkowski
  convolutional neural networks,'' in \emph{Proceedings of the IEEE Conference
  on Computer Vision and Pattern Recognition}, 2019, pp. 3075--3084.

\bibitem{tang2020searching}
H.~Tang, Z.~Liu, S.~Zhao, Y.~Lin, J.~Lin, H.~Wang, and S.~Han, ``Searching
  efficient 3d architectures with sparse point-voxel convolution,'' \emph{arXiv
  preprint arXiv:2007.16100}, 2020.

\bibitem{qi2016pointnet}
C.~R. Qi, H.~Su, K.~Mo, and L.~J. Guibas, ``Pointnet: Deep learning on point
  sets for 3d classification and segmentation,'' \emph{arXiv preprint
  arXiv:1612.00593}, 2016.

\bibitem{qi2017pointnetplusplus}
C.~R. Qi, L.~Yi, H.~Su, and L.~J. Guibas, ``Pointnet++: Deep hierarchical
  feature learning on point sets in a metric space,'' \emph{arXiv preprint
  arXiv:1706.02413}, 2017.

\bibitem{jiang2018pointsift}
M.~Jiang, Y.~Wu, T.~Zhao, Z.~Zhao, and C.~Lu, ``Pointsift: A sift-like network
  module for 3d point cloud semantic segmentation,'' \emph{arXiv preprint
  arXiv:1807.00652}, 2018.

\bibitem{zhang2019shellnet}
Z.~Zhang, B.-S. Hua, and S.-K. Yeung, ``Shellnet: Efficient point cloud
  convolutional neural networks using concentric shells statistics,'' in
  \emph{Proceedings of the IEEE International Conference on Computer Vision},
  2019, pp. 1607--1616.

\bibitem{li2018pointcnn}
Y.~Li, R.~Bu, M.~Sun, W.~Wu, X.~Di, and B.~Chen, ``Pointcnn: Convolution on
  x-transformed points,'' in \emph{Advances in Neural Information Processing
  Systems}, 2018, pp. 820--830.

\bibitem{thomas2019kpconv}
H.~Thomas, C.~R. Qi, J.-E. Deschaud, B.~Marcotegui, F.~Goulette, and L.~J.
  Guibas, ``Kpconv: Flexible and deformable convolution for point clouds,''
  \emph{arXiv preprint arXiv:1904.08889}, 2019.

\bibitem{kochanov2020kprnet}
D.~Kochanov, F.~K. Nejadasl, and O.~Booij, ``Kprnet: Improving projection-based
  lidar semantic segmentation,'' \emph{arXiv preprint arXiv:2007.12668}, 2020.

\bibitem{wang2019dynamic}
Y.~Wang, Y.~Sun, Z.~Liu, S.~E. Sarma, M.~M. Bronstein, and J.~M. Solomon,
  ``Dynamic graph cnn for learning on point clouds,'' \emph{Acm Transactions On
  Graphics (tog)}, vol.~38, no.~5, pp. 1--12, 2019.

\bibitem{voigtlaender2019feelvos}
P.~Voigtlaender, Y.~Chai, F.~Schroff, H.~Adam, B.~Leibe, and L.-C. Chen,
  ``Feelvos: Fast end-to-end embedding learning for video object
  segmentation,'' in \emph{Proceedings of the IEEE/CVF Conference on Computer
  Vision and Pattern Recognition}, 2019, pp. 9481--9490.

\bibitem{oh2019video}
S.~W. Oh, J.-Y. Lee, N.~Xu, and S.~J. Kim, ``Video object segmentation using
  space-time memory networks,'' in \emph{Proceedings of the IEEE International
  Conference on Computer Vision}, 2019, pp. 9226--9235.

\bibitem{zhou2019enhanced}
Z.~Zhou, L.~Ren, P.~Xiong, Y.~Ji, P.~Wang, H.~Fan, and S.~Liu, ``Enhanced
  memory network for video segmentation,'' in \emph{Proceedings of the IEEE
  International Conference on Computer Vision Workshops}, 2019, pp. 0--0.

\bibitem{zhou2019motion}
Q.~Zhou, Z.~Huang, L.~Huang, Y.~Gong, H.~Shen, W.~Liu, and X.~Wang,
  ``Motion-guided spatial time attention for video object segmentation,'' in
  \emph{Proceedings of the IEEE International Conference on Computer Vision
  Workshops}, 2019, pp. 0--0.

\bibitem{yang2020collaborative}
Z.~Yang, Y.~Wei, and Y.~Yang, ``Collaborative video object segmentation by
  foreground-background integration,'' \emph{arXiv preprint arXiv:2003.08333},
  2020.

\bibitem{tatarchenko2018tangent}
M.~Tatarchenko, J.~Park, V.~Koltun, and Q.-Y. Zhou, ``Tangent convolutions for
  dense prediction in 3d,'' in \emph{Proceedings of the IEEE Conference on
  Computer Vision and Pattern Recognition}, 2018, pp. 3887--3896.

\bibitem{hu2020randla}
Q.~Hu, B.~Yang, L.~Xie, S.~Rosa, Y.~Guo, Z.~Wang, N.~Trigoni, and A.~Markham,
  ``Randla-net: Efficient semantic segmentation of large-scale point clouds,''
  in \emph{Proceedings of the IEEE/CVF Conference on Computer Vision and
  Pattern Recognition}, 2020, pp. 11\,108--11\,117.

\bibitem{yan2020sparse}
X.~Yan, J.~Gao, J.~Li, R.~Zhang, Z.~Li, R.~Huang, and S.~Cui, ``Sparse single
  sweep lidar point cloud segmentation via learning contextual shape priors
  from scene completion,'' \emph{arXiv preprint arXiv:2012.03762}, 2020.

\bibitem{geiger2012we}
A.~Geiger, P.~Lenz, and R.~Urtasun, ``Are we ready for autonomous driving? the
  kitti vision benchmark suite,'' in \emph{2012 IEEE conference on computer
  vision and pattern recognition}.\hskip 1em plus 0.5em minus 0.4em\relax IEEE,
  2012, pp. 3354--3361.

\end{thebibliography}
}

\end{document}